\documentclass[runningheads]{llncs}
\usepackage{graphicx}

\usepackage{tikz}
\usepackage{comment}
\usepackage{amsmath,amssymb} 
\usepackage{color}

\usepackage{wrapfig}
\usepackage{multirow}
\usepackage{multicol}
\usepackage{array}

\begin{document}
\pagestyle{headings}
\mainmatter
\def\ECCVSubNumber{1045}  

\title{Deep Positional and Relational Feature Learning for Rotation-Invariant Point Cloud Analysis} 

\titlerunning{PR-invNet}
%
\author{Ruixuan Yu\inst{1,2}  \and
Xin Wei\inst{1}  \and
Federico Tombari\inst{2,3}  \and
Jian Sun\inst{1} }
\authorrunning{R. Yu et al.}
%
\institute{Xi'an Jiaotong University, China \and Technical University of Munich, Germany \and Google\\ 
\email {\{yuruixuan123, wxmath\}@stu.xjtu.edu.cn}, \\ tombari@google.com, jiansun@xjtu.edu.cn}

\maketitle

\begin{abstract}
In this paper we propose a rotation-invariant deep network for point clouds analysis. Point-based deep networks are commonly designed to recognize roughly aligned 3D shapes based on point coordinates, but suffer from performance drops with shape rotations. Some geometric features, e.g., distances and angles of points as inputs of network, are rotation-invariant but lose positional information of points. In this work, we propose a novel deep network for point clouds by incorporating positional information of points as inputs while yielding rotation-invariance. The network is hierarchical and relies on two modules: a positional feature embedding block and a relational feature embedding block.
Both modules and the whole network are proven to be rotation-invariant when processing point clouds as input. 
Experiments show state-of-the-art classification and segmentation performances on benchmark datasets, and ablation studies demonstrate effectiveness of the network design.
 
\keywords{Rotation-invariance, point cloud, deep feature learning.}
\end{abstract}

\section{Introduction}

Point clouds are widely employed as a popular 3D representation for objects and scenes. They are generated by most current acquisition techniques and 3D sensors, and used within well-studied application fields such as autonomous driving, archaeology, robotics, augmented reality, to name a few. Among these applications, shape recognition and segmentation are two fundamental tasks focusing on automatically recognizing and segmenting 3D objects or object parts  \cite{qi2017pointnet2,li2018pointcnn,pham2019jsis3d}.

The majority of 3D object recognition approaches are currently based on deep learning \cite{qi2017pointnet2,li2018pointcnn,su2015multi,esteves2019equivariant,wu20153d,qi2016volumetric}. Most of the point-based methods take positional information, such as point coordinates or normal vectors on aligned 3D objects, as inputs for the network, then learn deep features suitable for the task at hand. These methods now achieve state-of-the-art performance for 3D recognition. PointNet \cite{qi2017pointnet} firstly designed a network to process point clouds by taking point coordinates as inputs. Following works such as \cite{qi2017pointnet2,li2018pointcnn,xu2018spidercnn,wang2019dynamic,rao2019spherical} developed various convolution operations on point clouds which brought performance improvements. These advances justify that designing networks based on the positional information of 3D points on aligned object shapes is an effective way for shape recognition.

Nevertheless, in many scenarios such as, e.g., 6D pose estimation for robotics/AR \cite{Yimin2016A,Xiang2017PoseCNN} and CAD models \cite{HaraldTracking} in industrial environments, or analysis of molecules \cite{bero2014rotation,berenger2014a} where small scale objects are uncontrolled, a major limitation of above point-based methods is that they tend to be rotation-sensitive, and their performance drops dramatically when tested on shapes under arbitrary rotations in the 3D rotation group SO(3). A remedy for this is to develop robustness against rotations by augmenting training dataset with arbitrary rotations, e.g., the SO(3)/SO(3) mode in Spherical CNN~\cite{esteves2018learning}. However, it is not efficient to use data augmentation for learning rotation-invariance, since the enlarged dataset requires a large computational cost. Also, the developed robustness to rotations of the network will be up to the seen augmentations during training.

\begin{wrapfigure}{r}{0.55\textwidth}
\setlength{\abovecaptionskip}{0pt} 
\setlength{\belowcaptionskip}{10pt} 
\vspace{-15pt}
\includegraphics[width=0.55\textwidth]{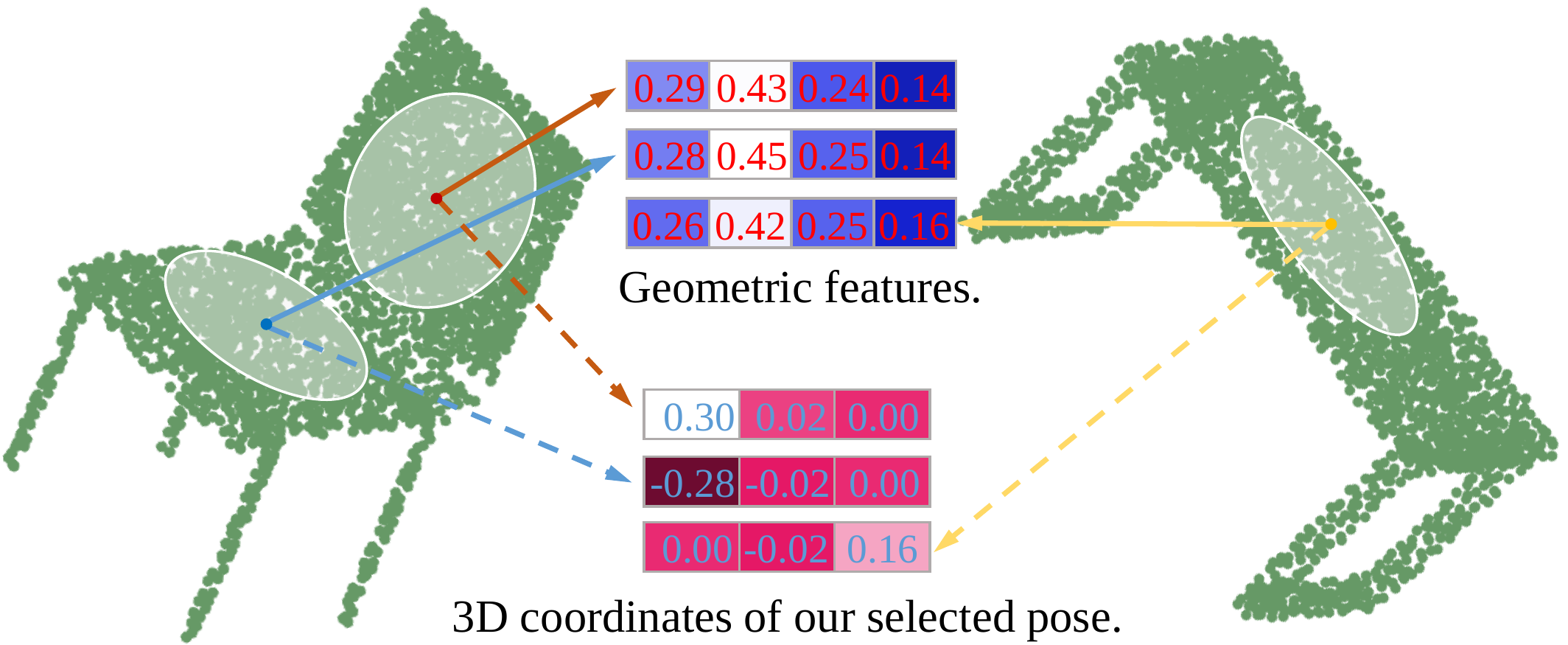}
\caption{The three circles highlight examples of local planar surfaces. Their center points are similar with geometric features~\cite{chen2019clusternet}, but distinguishable by point coordinates of selected pose as features in our approach. Features of each center point are shown in rows of tables.}
\label{fig:figure_front_page}
\vspace{-25pt}
\end{wrapfigure}
To deal with rotation-sensitivity, an alternative way is to represent shapes with geometric features such as distances and angles. Spherical CNN~\cite{esteves2018learning} and $a^3$SCNN~\cite{Liu2019DeepL3} are representative works that take distances and angles as spherical functions. They learn rotation-equivariant features by defining convolutions on spherical functions, and then aggregate features by global average pooling for rotation-invariance. Note that these representations are not rigorously rotation-invariant due to discretization on sphere. ClusterNet~\cite{chen2019clusternet} and RIConvNet~\cite{zhang2019rotation} also take rotation-invariant geometric features as inputs and design specific convolutions on local point cloudsan. However, directly transforming point coordinates to rotation-invariant geometric features may lose positional information of point cloud, which is essential to recognize 3D shapes when they are (roughly) aligned. For example, in Fig. \ref{fig:figure_front_page}, three local surfaces within the circles are flat surfaces belonging to different parts of two shapes. If we represent the center points of these local surfaces by geometric features (e.g., in ClusterNet~\cite{chen2019clusternet}) agnostic to their positional information, these features may not distinguish different parts of the same shape or different shapes. 

To achieve rotation-invariant shape recognition with high accuracy, we propose a novel deep architecture by learning positional and relational features of point cloud. It takes both point coordinates and geometric features as network inputs, and achieves rotation-invariance by designing a \textit{Positional Feature Embedding block (PFE-block)} and a \textit{Relational Feature Embedding block (RFE-block)}, both of which are rotation-invariant. For both shape classification and segmentation, invariant features on point cloud guarantee that classification label of shape and segmentation label of each point are invariant to shape rotation. 

It seems to be contradictory to learn rotation-invariant features with point coordinates as input. We design the PFE-block as composition of a pose expander, a pose selector and a positional feature extractor. The PFE-block is proven to be able to produce invariant features for points agnostic to shape rotations.  The pose expander maps the shape into a rotation-invariant pose space, then the pose selector selects a unique pose from the space, whose point coordinates are more  discriminative than geometric features as shown in Fig. \ref{fig:figure_front_page}. With this selected pose, we extract its positional features by a positional feature extractor. The RFE-block further enhances the deep feature representation of the point cloud with relational convolutions, where the weights are learned based on the relations of neighboring points. This block is also rotation-invariant.

As a summary, we propose two novel rotation-invariant network blocks and a deep hierarchical network jointly learning positional and relational features as shown in Fig. \ref{fig:pipleline}. Our network is one of the very few works that achieve rigorous rotation-invariance. For both point cloud shape classification and segmentation, we achieve state-of-the-art performances on commonly used benchmark datasets.

\section{Related Work}

\subsection{Point-based deep learning}
Point cloud is a basic representation for 3D shape, and point coordinates are taken as raw features by most point-based networks to carry out tasks on aligned shapes. PointNet~\cite{qi2017pointnet} is the first effort that takes point coordinates as raw features and embeds positional information of points to deep features followed by max-pooling to be a global shape descriptor. Afterwards, a series of works attempt to improve feature embedding by designing novel deep networks on point clouds. PointNet++~\cite{qi2017pointnet2} builds hierarchical architecture with PointNet as local encoder. More works such as SpiderCNN~\cite{xu2018spidercnn}, PointCNN~\cite{li2018pointcnn}, KCNet~\cite{shen2018mining}, PointConv \cite{wu2019pointconv}, RS-CNN~\cite{liu2019relation} learn point cloud features with various local convolutions starting from point coordinates or additional geometric features. Though these works have achieved state-of-the-art performance for various shape analysis tasks, they are sensitive to shape rotation. In our work, we build a novel network with building blocks being able to achieve rotation-invariance, but with positional raw features as inputs. Our network can explore and encode positional information of 3D shapes while precluding the disadvantage of rotation-sensitiveness.

\subsection{Rotation-robust representations}
There are several ways to improve rotation-robustness. Data augmentation by randomly rotating shapes in training set is widely applied in training 3D networks as in~\cite{qi2017pointnet,xu2018spidercnn}. Furthermore, transformer network~\cite{jaderberg2015spatial}  was generalized and utilized in 3D recognition \cite{qi2017pointnet,bas20173d,mukhaimar2019pl}. The transformer implicitly learns alignments among shapes either in 3D space or feature space which may reduce the impact of shape rotations. As shown in~\cite{chen2019clusternet}, these techniques commonly improved robustness but can not achieve rigorous invariance to rotations.

Rotation-invariant geometric features can be taken as network inputs. Traditionally, these hand-crafted features are widely utilized to represent 3D shapes in \cite{stein1992structural,sun2001surface,zhong2009intrinsic,rusu2009fast,tombari2010unique}, and further introduced as inputs of deep networks~\cite{chen2019clusternet,zhang2019rotation,deng2018ppf}. Most of them utilize distances and angles \cite{zhang2019rotation,deng2018ppf} and sine / cosin values of the angles \cite{chen2019clusternet} as raw point-wise features. They encode relations among points and achieve invariance to arbitrary shape rotations. However, the geometric features may lose discriminative clues that are contained in point positions / orientations.

Another way to achieve rotation-robustness is to learn rotation-equivariant features. \cite{thomas2018tensor} built filters on spherical harmonics and learned locally equivariant features to 3D rotations and translations. Group convolution~\cite{cohen2016group} is a generalized convolution to achieve rotation-equivariant feature learning based on rotation-group. This idea is extended to 3D domain \cite{worrall2018cubenet} with various discretized rotation-groups. Spherical CNN \cite{esteves2018learning} and $a^3$SCNN \cite{Liu2019DeepL3} defined group convolution on sphere  which can be taken as an approximation to infinite group SO(3). They also require the input of the networks to be rotation-equivariant if regarding features as functions over shapes. Commonly, these methods lead to rotation-robustness by global aggregation such as average-pooling or max-pooling, but rigorous rotation-invariance is not ensured due to the discretized rotation groups.

\begin{figure}[t]
\begin{center}
\includegraphics[width=1\linewidth]{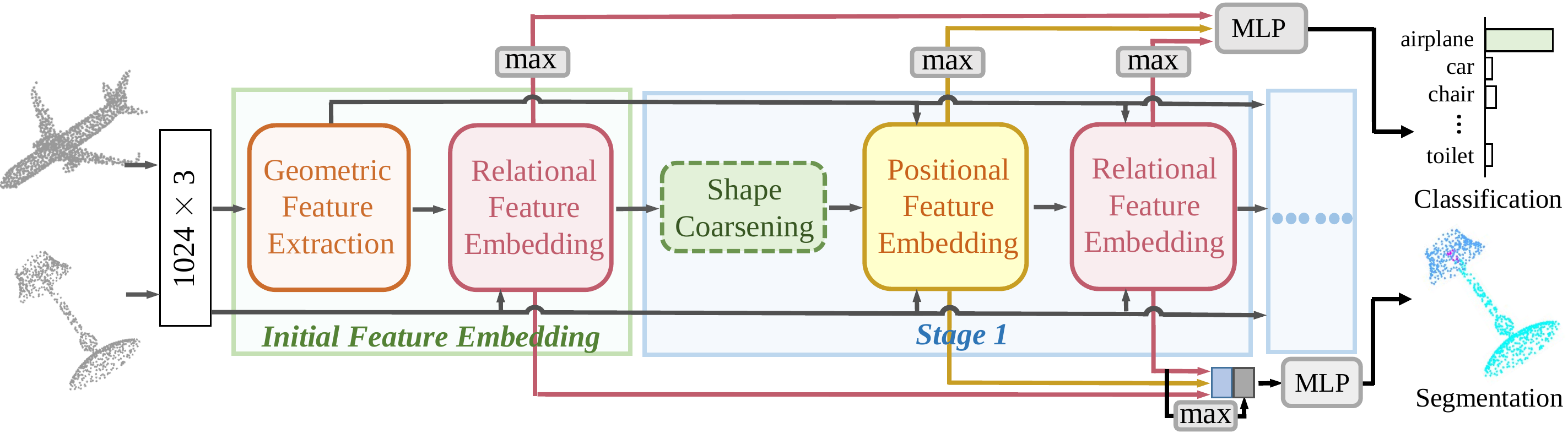}
\end{center}
\caption{PR-invNet consists of geometric feature extraction, positional feature embedding block, relational feature embedding block. PR-invNet is designed for classification and segmentation. Shape coarsening is not utilized in segmentation. }
\label{fig:pipleline}
\end{figure}

Instead of requiring rotation-invariant geometric features as inputs, our network can directly learn positional features from point coordinates as inputs while achieving rigorous rotation-invariance based on PFE-block. It embeds the input shape into a rotation-invariant pose space, from which we derive a discriminative pose by pose selector to be sent to positional feature extractor, and this operation is rotation-invariant and justified to be effective in experiments.

\section{Method}

In this section, we introduce our rotation-invariant network, dubbed \textit{PR-InvNet}. It aims at deep positional and relational feature learning, based on, respectively, the PFE-block and RFE-block. 
As in Fig. \ref{fig:pipleline}, it consists of several stages. The initial feature embedding stage is composed of geometric feature extraction and RFE-block. 
The output features of the PFE and RFE blocks are max-pooled over point cloud and concatenated, then fed to a MLP for shape classification. For point cloud segmentation, we switch off shape coarsening operation, and the concatenated features of PFE-blocks and RFE-blocks are further concatenated with globally max-pooled features before being fed to MLP for point label prediction. Each stage is now detailed in the following.

\subsection{Geometric feature extraction} \label{geo_fea_section}
\begin{figure}[t]
\begin{center}
\includegraphics[width=0.9\linewidth,height=0.8in]{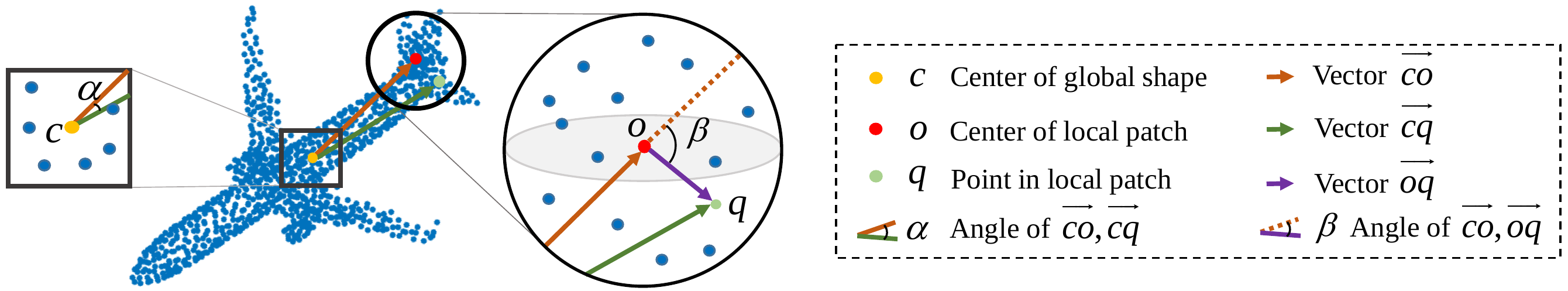}
\end{center}
\caption{Illustration of geometric feature extraction for point $o$ which is based on distances and angles. See Sect. 3.1 for details.}
\label{fig:geometric}
\end{figure}

PR-invNet takes geometric features similar to \cite{chen2019clusternet} as additional raw input features complementary to point coordinates\footnote{Even without geometric features, our network would be still rotation-invariant.}. As shown in Fig. \ref{fig:geometric}, $c$ and $o$ are the centers of the global shape and a local patch respectively. To compute geometric features of point $o$, we first consider its neighboring points. The feature of each of its neighboring point $q$ is computed as $[|\overrightarrow{cq}|,|\overrightarrow{oq}|, cos(\alpha), cos(\beta)]$,  with $[\cdot]$ being the concatenation operation. Then we concatenate features of all neighboring points together with the distance of $|\overrightarrow{co}|$ as the geometric feature of point $o$, i.e., 
\begin{equation}
g_o = [|\overrightarrow{co}|,\{|\overrightarrow{cq}|,|\overrightarrow{oq}|, cos(\alpha), cos(\beta)\}_{q \in N(o)}],
\end{equation}
where $N(o)$ denotes set of neighboring points of $o$.
For simplicity, we denote it  as $\{g_i\}_{i=1}^N$ with $N$ as number of points. It can be noted that $g_i$ is rotation-invariant, please refer to supplemental material for proof.
In our work, for each given point, the above geometric features are extracted and concatenated over a multi-scale neighborhoods (three neighborhoods determined by Euclidean distance with 8, 16, 32 points), and 8 neighboring points are uniformly sampled at each scale. Then the geometric feature of each point is in length of $3\times4\times8+1 = 97$.

\subsection{Positional feature embedding block}

The PFE-block aims to perform rotation-invariant feature embedding by incorporating positional information encoded in point coordinates.  
Since point coordinates are sensitive to shape rotations, one solution is to take all rotated versions (i.e., poses) of a shape into consideration, and fuse extracted features from them by global pooling. However, this is inefficient since there are infinite number of rotations in SO(3) to ensure rigorous rotation-invariance. We propose an idea that first maps input shape to a rotation-invariant pose space by \textit{pose expander}, then selects a  representative pose by \textit{pose selector}. Point coordinates of selected pose are utilized to extract positional features by \textit{feature extractor}.

\subsubsection{Pose expander.} The pose expander aims to map a shape to produce a rotation-invariant pose space.
Given a shape with point cloud $P\in  \mathcal{R}^{N\times 3}$, it is obvious that the pose space $\{\mathcal{T}_{g}P\}_{g \in SO(3)}$ containing all possible rotated shapes in SO(3) is rotation-invariant. 
We aim to derive a compact pose space of input shape guaranteeing rotation-invariance. We achieve this by first normalizing the pose of each shape via Principal Components Analysis (PCA), then applying a discretized rotation-group in SO(3) to the normalized shape to derive a pose space.

\begin{wrapfigure}{r}{0.4\textwidth}
\vspace{-20pt}
\includegraphics[width=0.4\textwidth]{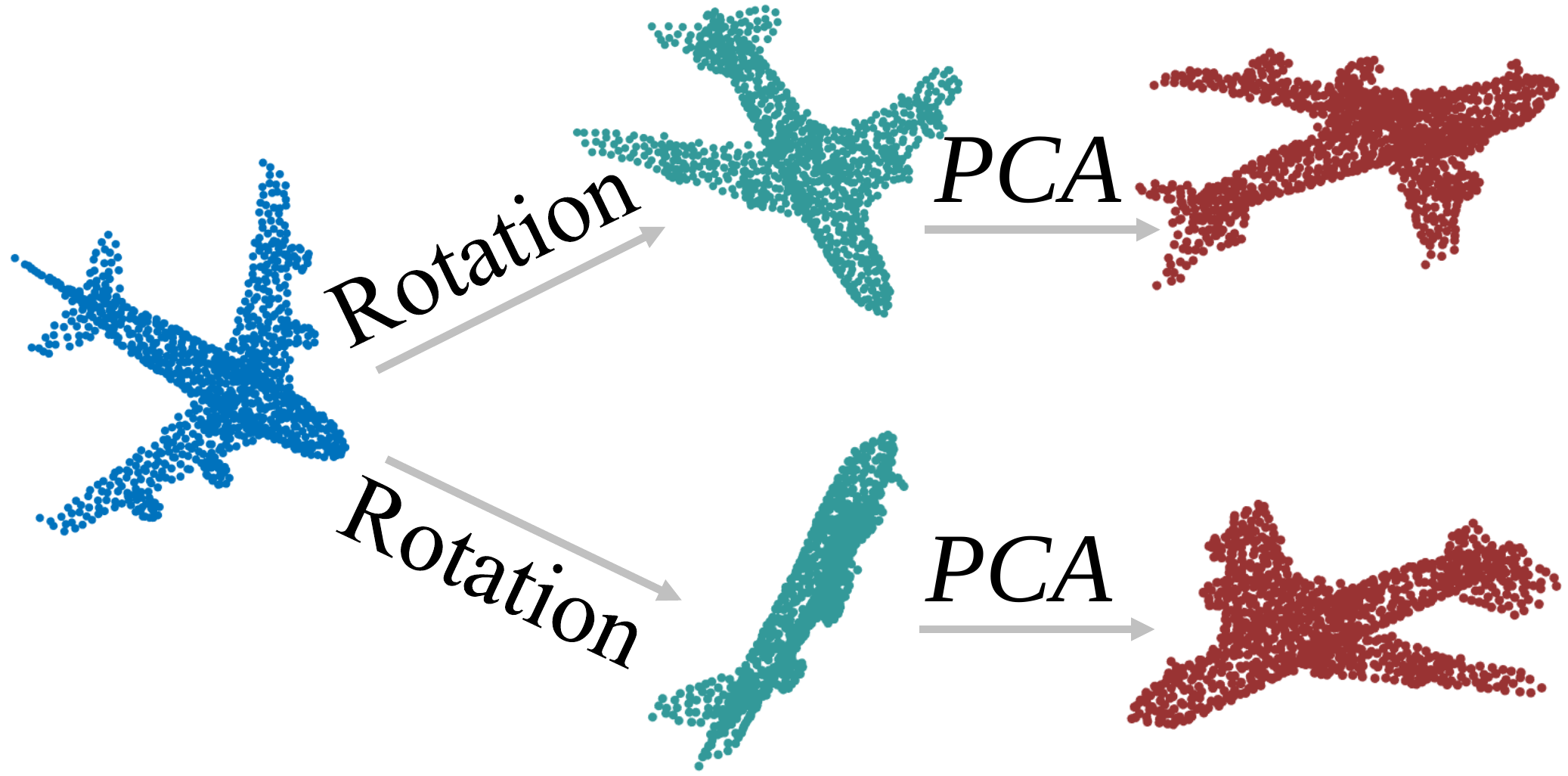}
\caption{Two rotated versions of one shape may result in two symmetric and different poses by PCA.}
\label{fig:figure_eigen_shape}
\vspace{-20pt}
\end{wrapfigure}

We first apply PCA on the point coordinates of a shape to normalize its pose using a coordinate system composed of the three eigen-vectors. We indicate the normalized shape as $\widehat{P}$. Note that pose normalization by PCA is not injective. For example, eigen-decomposition may result in different signs of eigen-vectors, therefore results in different normalized shapes. In Fig. \ref{fig:figure_eigen_shape}, a shape with two rotated versions may produce two different normalized shapes, which are symmetric.

To solve the ambiguity of PCA-normalized shape, we enumerate all possible signs of eigen-vectors, and construct a PCA-normalized pose space with eight normalized shapes based on a symmetric group $G_S$
\begin{equation}
G_S = \{g_{s} |g_{s} = \left[ \begin{array}{ccc}
x & ~0 & ~0\\
0& ~y & ~0\\
0 & ~0 & ~z
\end{array}
\right], x,y,z\in \{-1,1\}, s = 1,2,\cdots,8\},
\end{equation}
Then the PCA-normalized pose space is obtained as
\begin{equation}
H_S(P) = \{\widehat{P}g_s|g_s \in G_S \},
\label{eqn:h_s_p}
\end{equation}
where $H_S(P)$ contains eight possible PCA-normalized shapes corresponding to different possible signs of eigen-vectors, and this pose space is denoted as \textit{sym-space}. Theorem 1 proves that $H_S(P)$ is rotation-invariant for input shape $P$. 
 
Though sym-space is rotation-invariant, PCA may not always well align the poses of different shapes even from the same category. We further expand the pose space from $H_S(P)$ to increase the possibility of containing aligned poses for different shapes. By discretizing SO(3) using limited representative rotations, we enlarge sym-space $H_S(P)$ to \textit{rot-space} $H_{SR}(P)$ using rotation group $G_R$ as
\begin{equation}
H_{SR}(P) = \{\widehat{P}g_s g_r| g_s \in G_S, g_r \in G_R\},
\label{eq:rotSpace}
\end{equation}
where $G_R$ can be any discretized subgroup of SO(3).  $H_{SR}(P)$  is also rotation-invariant for $P$ as proved by Theorem 1. Compared with $H_{S}(P)$, rot-space $H_{SR}(P)$ contains more poses with higher chances to include the same or similar poses of different shapes. In our implementation, we use rotation group $A_5$ , i.e., alternating group of a regular dodecahedron (or a regular icosahedron) \cite{zimmermann2011finite} to construct the rot-space. By deleting duplicate shapes, we finally map the shape with rotation set  $A_5\times Z_2$ into rot-space with 120 poses.  

\subsubsection{Pose selector.}
Given a shape $P$, we have derived a rotation-invariant pose space $H_{SR}(P)$, i.e., the rot-space. One naive way is to extract features from different poses of the shape in pose space followed by orderless operations, e.g., average- or max-pooling, to aggregate features as a rotation-invariant representation for shape $P$. However, it is computationally intensive to extract features from multiple poses of shape.
We design a simple but effective method by selecting the most representative shape pose using a pose selector, and learn positional features from this single selected shape instead than from all shapes in pose space. 

The pose selector is designed as a \textit{multi-head neural network} $\Psi(\cdot)$ over shape poses in $H_{SR}(P)$. We score each pose by
$
v_{s}=\Psi(\bar{P}_{s})$ for pose $ \bar{P}_{s}  \in H_{SR}(P),  \ s=1,\cdots,|H_{SR}(P)|
$.
$v_{s}\in  \mathcal{R}^{N_h}$ is a vector generated by $N_h$ heads of $\Psi(\cdot)$. In our implementation, $\Psi(\cdot)$ is designed as a simple network with a point-wise fully connected (FC) layer, a max-pooling layer over all points, followed by an additional FC layer and softmax with $N_h$ output scores for a point cloud. Similar to multi-head attention~\cite{vaswani2017attention}, the multiple elements in vector $v_{s}$ reflect the responses of $\bar{P}_{s}$ to different modes of the neural network. 
For a shape pose $ \bar{P}_{s}$, its final score is set as the largest response in vector $v_{s}$, i.e., 
$\bar{v}_{s} = \text{max}\{v_{s}\}$,
and $\bar{v}_{s}$ is a scalar. Then the selected representative shape $\widetilde{P}$ from pose space $H_{SR}(P)$ for shape $P$ is the shape pose with the largest score.

Theorem 1 proves that the sym-space, rot-space  and the selected pose are rotation-invariant to $P$ (please see supplementary material for proof).

\begin{theorem}
Denoting $P\in  \mathcal{R}^{N\times 3}$ as point cloud of a shape, its sym-space $H_S(P)$ in Eqn. (\ref{eqn:h_s_p})  and rot-space $H_{SR}(P)$ in Eqn. (\ref{eq:rotSpace}) are rotation-invariant for $P$, i.e., for rotated shape $Q = PR$ of $P$ with any rotation matrix $R\in  \mathcal{R}^{3\times 3}$, we have $H_S(P) = H_S(Q)$ and $H_{SR}(P) = H_{SR}(Q)$. Assuming the pose selector is injective, the selected pose of $P$ is rotation-invariant, i.e., $\widetilde{P}=\widetilde{Q}$.
\end{theorem}

\begin{figure}[t]
\begin{center}
\includegraphics[width=1\linewidth]{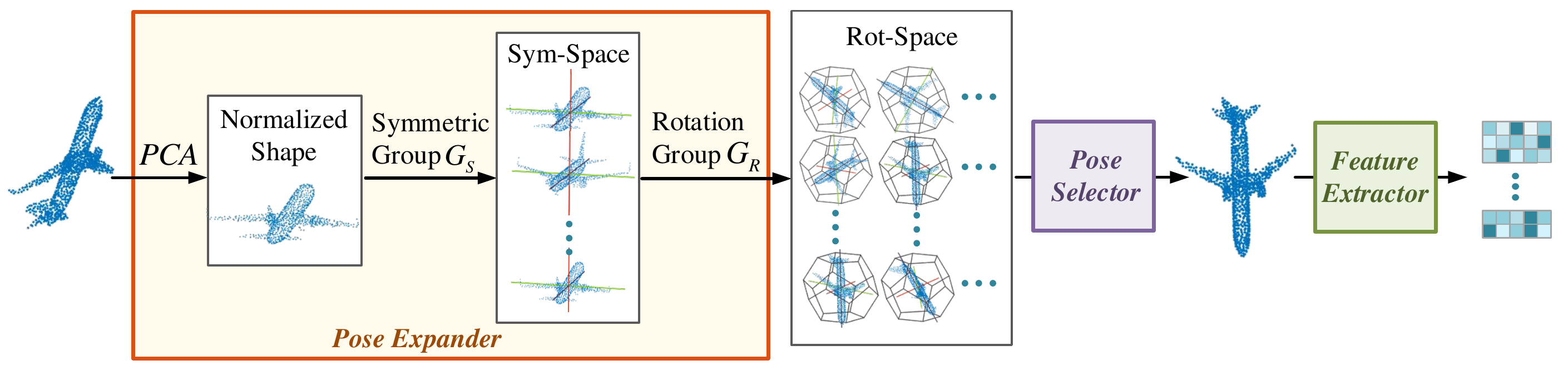}
\end{center}
\caption{Overview of the PFE-block. It maps a shape to be rotation-invariant pose space, i.e., rot-space, then selects a discriminative pose via pose selector, followed by feature extractor. This block is rotation-invariant.}
\label{fig:pfe_block}
\end{figure}

\subsubsection{Feature extractor.}
With the selected shape pose $\widetilde{P} \in  \mathcal{R}^{N\times3}$ of input shape $P$, we can extract its point-wise features based on the point coordinates like the traditional point-based networks, e.g., \cite{qi2017pointnet2,li2018pointcnn,xu2018spidercnn}.  
 Specifically, given point-wise coordinates $\tilde{p}_i$, geometric feature $g_i$ (extracted from input shape $P$), and already extracted point features $f_i$, we concatenate $[g_i,f_i,\tilde{p}_i]$ as point-wise input features for our feature extractor. Motivated by the structure of PointNet++\cite{qi2017pointnet2}, for each point $i$, its point feature is extracted as
\begin{equation}
f'_i =  \Gamma\left(\{[g_j,f_j,\tilde{p}_j]\}_{j \in N(i)}\right)
\label{eqn:feat_extr}
\end{equation}
where $ N(i)$ is the local neighborhood of point $i$ with $K$ points including point $i$. $\Gamma(\cdot)$ is the feature extractor, which is designed as a shared MLP for points, followed by max-pooling over neighborhood. 
Fig. \ref{fig:pfe_block} shows the pipeline of the PFE-block. Since the selected pose is rotation-invariant to input shape by Theorem 1, its point coordinates $\tilde{p}_i$ are also rotation-invariant. Hence, features in Eqn.~(\ref{eqn:feat_extr}) are rotation-invariant because all inputs of $\Gamma$ are rotation-invariant.

\subsection{Relational feature embedding block}
The relational feature embedding block (RFE-block) is specifically designed to explore and learn point features by modeling interactions of neighboring points.
For each point $i$, we aggregate the features of its local neighborhood with $K$ points including point $i$ by convolution. Assuming the neighboring points are indexed by $k$ sorted by increasing Euclidean distance to point $i$, and corresponding point features are $\{f_k \in  \mathcal{R}^{d}\}_{k=1}^{K}$, the updated feature $f'_i \in  \mathcal{R}^{d'}$ for point $i$ is
\begin{equation}
f'_{il} = \sum_{k \in N(i)}\sum_{j=1}^{d} f_{kj}W_{kjl},  \mbox{ for } l = 1,\cdots,d'
\label{eqn:conv}
\end{equation}
where $f_{kj}$ is the $j$-th feature channel of the $k$-th neighboring point, $f'_{il}$ is the $l$-th channel of the updated feature $f_i'$. $W_{kjl}$ is defined as
\begin{equation}
\begin{split}
& W_{kjl} =\Phi([C_k^{p},C_k^{g},C_k^{f},g_k])_{j}\Theta_{kjl}, 
\end{split}
\label{eqn:relation}
\end{equation}
where $C_k^p,C_k^g,C_k^f \in  \mathcal{R}^K$ are $K$-d vectors denoting correlations of centered point coordinates, geometric features, point features respectively between $k$-th point and all $K$ points in local neighborhood, followed by $l_2$-normalization. For example, for centered point coordinates, i.e., coordinates subtracted by that of center pixel $i$, 
$C_k^p = \text{Normalize}([p_k^{\top}p_1,p_k^{\top}p_2,\cdots,p_k^{\top}p_K])$, where we use $p$ to represent the centered point coordinates.
$C_k^g,C_k^f$ are similarly defined as inner products of corresponding features.
$\Phi(\cdot)$ is modeled as MLP, embedding relationships of neighboring points to convolution weight with learnable parameters $\Theta_{kjl}$. 
It is easy to verify that weights in Eqn.~(\ref{eqn:relation}) are rotation-invariant because the inputs of $\Phi$ are rotation-invariant defined by inner products.

\subsection{Network architecture}

PR-invNet is designed by concatenating rotation-invariant PFE-block and RFE-block. As shown in Fig. \ref{fig:pipleline}, given a shape, we first initialize the point-wise features by initial feature embedding, followed by three stages concatenating PFE-block and RFE-block to extract positional and relational features. For classification, we use an additional shape coarsening operation at each stage to construct a hierarchy of coarsened points, implemented by farthest point sampling with sampling rate of 1/2. 
The intermediate features generated by PFE-blocks and RFE-blocks of different stages are concatenated and aggregated by a MLP set as two layers of FC+BN+ReLU+Dropout, followed by a FC to output confidence for shape categorization or  segmentation. The number of hidden units of the first two FCs are 512, 256, and dropout probability is 0.5. The number of neighboring points in Eqns. (\ref{eqn:feat_extr}-\ref{eqn:conv}) is 16.
For an input shape of PR-invNet, we conduct PCA to normalize shape pose only one time, and all PFE-blocks in different stages derive rot-space by Eqn.(\ref{eq:rotSpace}) starting from point cloud of this normalized pose or its coarsened point clouds. 
More details of pose selector $\Psi$, feature extractor $\Gamma$, convolution filters $\Phi$, $\Theta$ in RFE-block are introduced in  supplementary material.

\section{Experimental results}
In this section we evaluate and compare with state-of-the-art methods for rotated shape classification and segmentation on the following 3D datasets.

\textbf{ModelNet40 \cite{wu20153d}} consists of 12,311  shapes from
40 categories, with 9,843 training and 2,468 test objects for shape classification. 

\textbf{ShapenetCore55 \cite{savva2016shrec16}} contains two subsets: `normal', `perturbed', with aligned and randomly rotated shapes in $SO(3)$ respectively. 51,190 models are categorized into 55 classes with training, validation, test sets in ratios of 70\%, 10\%, 20\%. We uniformly sample points from the original mesh data for both subsets.

\textbf{ShapeNet part \cite{chang2015shapenet}} contains 16,880 shapes from 16 category of shapes, with 14,006 / 2,874 shapes for training / test, annotated with 50 parts. 

Each shape is represented by 1024 points. To train our PR-invNet, we take Adam optimizer with initial learning rate, epoch number as 0.001, 250, and the learning rate is exponentially decayed with decay rate and decay step as 0.7, 200000. The batch size for shape classification and segmentation are 16 and 8 respectively. It takes 286.6ms, 672.1ms in one iteration for shape classification and segmentation respectively on a NVIDIA 1080 Ti GPU.

\subsection{3D shape classification}
We mainly compare with point-based methods and rotation-robust methods.

\begin{table}[h]
\caption{Shape classification accuracy on ModelNet40 (in $\%$).}
\begin{center}
\begin{tabular}{p{1.5cm} |l  p{1.6cm}<{\centering}  p{1.6cm}<{\centering} p{1.6cm}<{\centering} p{1.6cm}<{\centering}}
\hline
&~Method & Input size  & z/z  & SO(3)/SO(3) & z/SO(3)\\
\hline
\multirow{10}{1cm}{Rotation-sensitive}
&~VoxNet \cite{maturana2015voxnet} & $30^3$ & 83.0 &  87.3  & - \\
&~SubVolSup \cite{qi2016volumetric}  & $30^3$ & 88.5   & 82.7 & 36.6\\
&~SubVolSup MO \cite{qi2016volumetric} & $30^3$ & 89.5 & 85.0&45.5  \\
&~MVCNN 12x \cite{su2015multi} & $12 \times 224^2$ & 89.5  & 77.6 & 70.1\\
&~MVCNN 80x \cite{su2015multi}  & $80 \times 224^2$ & 90.2  & 86.0& \textbf{81.5}\\
&~PointNet \cite{qi2017pointnet}& $1024 \times 3$ &  87.0 &80.3 & 12.8\\
&~PointNet++ \cite{qi2017pointnet2} & $1024 \times 3$ & 89.3 & 85.0 & 28.6  \\
&~PointCNN  \cite{li2018pointcnn} & $1024 \times 3$ & 91.3 & 84.5 & 41.2\\
&~DGCNN  \cite{wang2019dynamic}    & $1024 \times 3$ & \textbf{92.2} &   \textbf{81.1} &20.6  \\
\hline
\multirow{5}{1cm}{Rotation-robust}
&~Spherical CNN \cite{esteves2018learning} & $2 \times64 \times 64$ & 88.9&76.9 &86.9 \\
&~$a^3$SCNN \cite{Liu2019DeepL3} & $2\times 165 \times 65$ & \textbf{89.6} & 87.9 & 88.7\\
&~RIConvNet \cite{zhang2019rotation} & $1024 \times 3$ &86.5 &86.4 &86.4 \\
&~ClusterNet \cite{chen2019clusternet} & $1024 \times 3$ &87.1 &87.1 &87.1 \\
&~RRI-PointNet++ & $1024 \times 3$ & 79.4&79.4 &79.4 \\
&~Proposed & $1024 \times 3$ &  89.2 &\textbf{89.2} & \textbf{89.2}\\
\hline
\end{tabular}
\end{center}
\label{tab:modelnet40}
\end{table}

\textbf{\textit{ModelNet40.}} As done in Spherical CNN~\cite{esteves2018learning}, we compare in the following three modes. (1) Both training and test on data augmented by azimuthal rotation (z/z); (2) Training with azimuthal rotation and test with arbitrary rotation
(z/SO(3)); (3) Both training and test with arbitrary rotation augmented data
(SO(3)/SO(3)). The results are presented in Table \ref{tab:modelnet40}. 
The traditional point-based methods, including PointNet \cite{qi2017pointnet}, PointNet++ \cite{qi2017pointnet2}, PointCNN \cite{li2018pointcnn}, DGCNN \cite{wang2019dynamic}, perform best in z/z mode, but decline sharply when testing on data augmented with arbitrary rotations (z/SO(3)). Though using arbitrarily augmented training data in SO(3)/SO(3) mode, their performances still drop with large gaps. For these rotation-robust methods including Spherical CNN  \cite{esteves2018learning}, $a^3$SCNN \cite{Liu2019DeepL3}, RIConvNet \cite{zhang2019rotation}, ClusterNet \cite{chen2019clusternet} and RRI\footnote{RRI is an essential part of ClusterNet, and the codes are provided by the authors.}-PointNet++, they are more robust in different modes. Considering that Spherical CNN and $a^3$SCNN rely on discretized angles in the sphere, they are not rigorously rotation-invariant and performance drops in modes (2)-(3). ClusterNet, RIConvNet and RRI-PointNet++ are rotation-invariant relying on input geometric features. Our PR-invNet is also rotation-invariant, and it achieves highest performance in modes (2-3) and second highest performance in mode (1) than all the rotation-robust methods, demonstrating its superiority for arbitrarily rotated shape classification.

\begin{table}[htbp]
\caption{Shape classification accuracy on ShapeNetCore55 (in $\%$). All the results of compared methods are achieved by running their code.}
\begin{center}
\begin{tabular}{p{1.3cm}| l  p{1.5cm}<{\centering}  p{1.5cm}<{\centering} p{1.5cm}<{\centering}}
\hline
&~Method           & Input size  &   Aligned & Perturbed\\
\hline
\multirow{5}{1cm}{Rotation-sensitive}
&~PointNet \cite{qi2017pointnet}      & 1024$\times$3 & 83.4 &74.6\\
&~PointNet++ \cite{qi2017pointnet2}   & 1024$\times$3 & 84.2 & 70.9\\
&~DGCNN  \cite{wang2019dynamic}      &  1024$\times$3   &  \textbf{86.3} & \textbf{74.1} \\
&~RS-CNN \cite{liu2019relation} & 1024$\times$3 & 85.9& 73.4\\
&~SpiderCNN \cite{xu2018spidercnn} & 1024$\times$3 & 79.8& 64.4\\
\hline
\multirow{4}{1cm}{Rotation-robust}
&~Spherical CNN \cite{esteves2018learning} & 2$\times64^2$ & 76.2 & 73.8\\
&~RIConvNet \cite{zhang2019rotation} & 1024$\times$3 & 78.5 & 76.9\\
&~RRI-PointNet++ & 1024$\times$3 & 70.8 & 67.9\\
&~Proposed & 1024$\times$3 & \textbf{78.9} & \textbf{77.6}\\
\hline
\end{tabular}
\end{center}
\label{tab:shapenetcore55}
\end{table}

\textbf{\textit{ShapeNetCore55.}}  We compare with point-based methods including PointNet \cite{qi2017pointnet}, PointNet++ \cite{qi2017pointnet2}, DGCNN \cite{wang2019dynamic}, RS-CNN \cite{liu2019relation}, SpiderCNN \cite{xu2018spidercnn}, and also rotation-robust Spherical CNN~\cite{esteves2018learning} and rotation-invariant RIConvNet \cite{zhang2019rotation}, RRI-PointNet++. The results are presented in Table \ref{tab:shapenetcore55}.
As shown in Table \ref{tab:shapenetcore55}, we achieve best performance in both aligned and perturbed datasets, compared with Spherical CNN, RIConvNet and RRI-PointNet++. Among all methods, we achieve highest accuracy on the perturbed dataset.

\subsection{3D shape segmentation}

\begin{table}[t]
\caption{Shape segmentation mIoU on ShapeNet part segmentation dataset (in $\%$).}
\begin{center}
\begin{tabular}{p{1.5cm}| p{2.7cm} p{2.3cm} p{1.4cm}  p{2.0cm}<{\centering} p{2.0cm}<{\centering}}
\hline
&~Method   & Input &  z/z & SO(3)/SO(3) & z/SO(3)\\
\hline
\multirow{5}{1cm}{Rotation-sensitive}
&~PointNet \cite{qi2017pointnet}      & xyz &76.2& 74.4 &37.8\\
&~PointNet++  \cite{qi2017pointnet2}  & xyz+normal &80.7 & \textbf{76.7} & \textbf{48.2}\\
&~PointCNN  \cite{li2018pointcnn}     & xyz &81.5 &71.4  &34.7 \\
&~DGCNN   \cite{wang2019dynamic}     &  xyz &78.8 &  73.3& 37.4\\
&~SpiderCNN \cite{xu2018spidercnn} & xyz+normal & \textbf{81.8} & 72.3& 42.9\\
\hline
\multirow{2}{1cm}{Rotation-robust}
&~RIConvNet \cite{zhang2019rotation} & xyz &75.6 & 75.5& 75.3\\
&~Proposed & xyz & \textbf{79.4} & \textbf{79.4} &\textbf{79.4}\\
\hline
\end{tabular}
\end{center}
\label{tab:shape_segmentation_iou}
\end{table}

\begin{wrapfigure}{r}{0.5\textwidth}
\vspace{-20pt}
\includegraphics[width=0.5\textwidth]{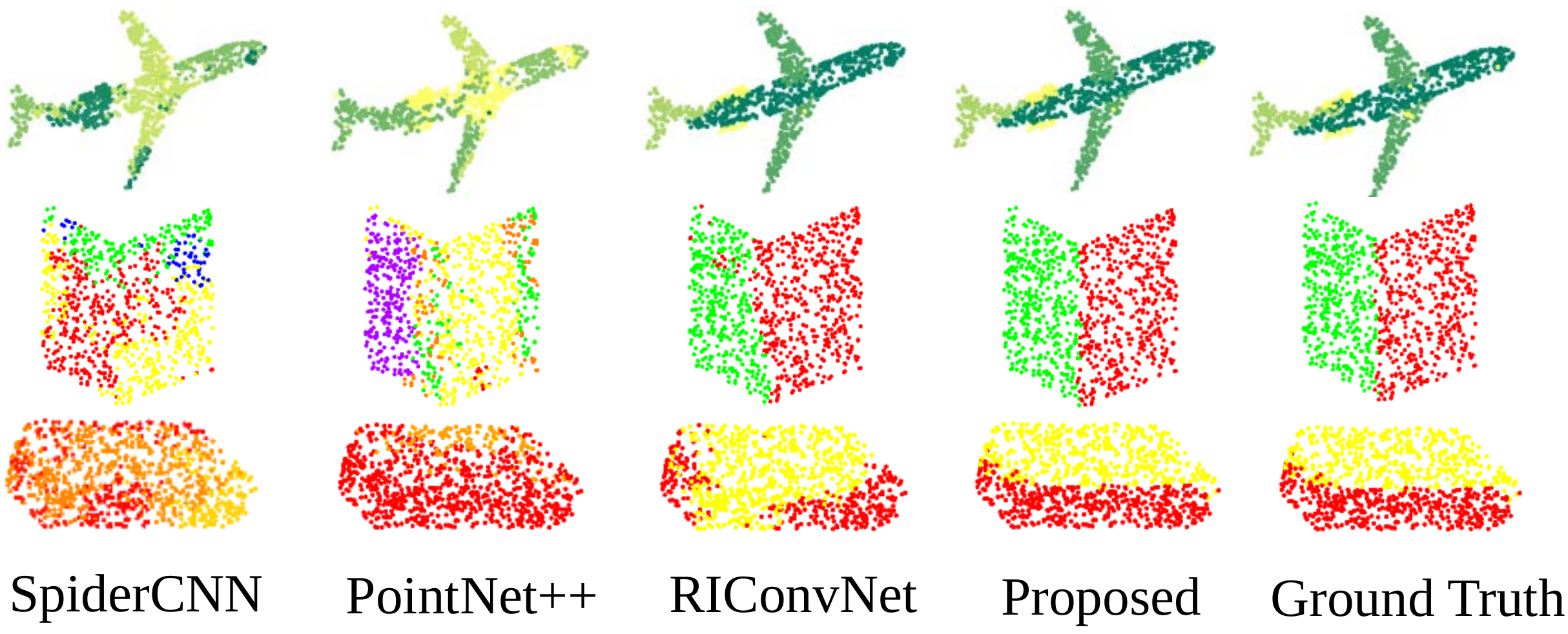}
\caption{Shape segmentation results. }
\label{fig:figure_show_seg}
\vspace{-15pt}
\end{wrapfigure}
We next apply PR-invNet to point cloud segmentation to predict the part label of each point. We use the same experimental settings as RIConvNet~\cite{zhang2019rotation}, and mean per-class IoU (mIoU, $\%$) are presented in Table \ref{tab:shape_segmentation_iou}. Compared with all rotation-sensitive point-based methods and rotation-robust methods, our PR-invNet outperforms them significantly in both SO(3)/SO(3) and z/SO(3) modes, i.e., more than $2.7\%$ and  $4.1\%$ mIoU improvements. We also report the per-class IoU for above settings in z/SO(3), SO(3)/SO(3) modes respectively in the supplemental material, and our PR-invNet achieves better performance than the rotation-robust RIConvNet on all categories.
In Fig. \ref{fig:figure_show_seg}, we show the segmentation results in z/SO(3) mode of several objects as well as the corresponding ground truth labels, and our predictions labels are reasonable and close to ground truth.

\makeatletter\def\@captype{table}\makeatother
\begin{minipage}{.65\textwidth}
\centering
\caption{Classification accuracy on ModelNet40 (\%).}
\vspace{10pt}
\label{tab:every_operation}
\begin{tabular}{l  p{1.0cm}<{\centering}  p{1.0cm}<{\centering} p{1.0cm}<{\centering} p{1.0cm}<{\centering}}
\hline
\multirow{2}{*}{~Method}     & Geo-fea  &   PFE-block & RFE-block & \multirow{2}{*}{Acc.}\\
\hline
~PR-invNet-noGeo & $\times$&$\surd$ &$\surd$ & 88.2\\
~PR-invNet-noPFE & $\surd$& $\times$&$\surd$ & 87.8\\
~PR-invNet-noRFE & $\surd$&$\surd$ & $\times$& 88.0\\
~PR-invNet &$\surd$ &$\surd$ &$\surd$ & \textbf{89.2}\\
\hline
\end{tabular}
\end{minipage}
\makeatletter\def\@captype{figure}\makeatother
\begin{minipage}{0.32\textwidth}
\centering
\includegraphics[width=3.3cm]{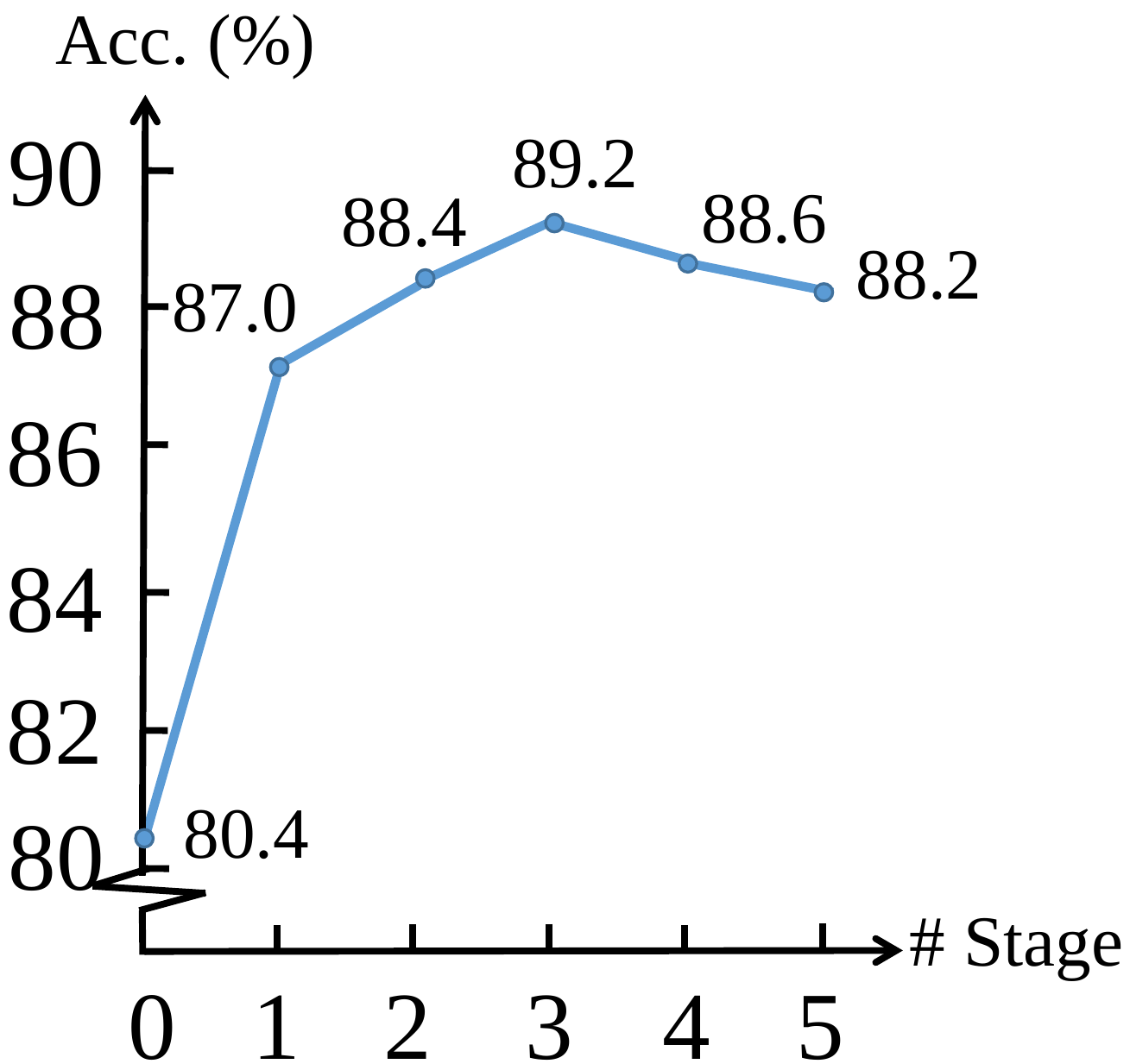}
\caption{Stage numbers.}
\label{fig:different_hierarchy}
\end{minipage}

\subsection{Ablation study}
In this section, we conduct ablation study on ModelNet40 to justify the effects of our network design, and also test the robustness of PR-invNet to noises.

\textit{\textbf{Effect of proposed blocks.}}
To evaluate the effectiveness of geometric feature extraction, PFE-block, and RFE-block, we conduct experiments that utilize networks without above blocks respectively, i.e., PR-invNet-noGeo, PR-invNet-noPFE, PR-invNet-noRFE, on ModelNet40, and present the experiment results in Table \ref{tab:every_operation}. Compared with PR-invNet, the networks of PR-invNet-noGeo, PR-invNet-noPFE, PR-invNet-noRFE achieve lower performance, demonstrating the need for each component. Note that all networks of above variants are also rigorously rotation-invariant, and they also achieve better performance than most of the rotation-sensitive and rotation-robust methods whose results are presented in Table \ref{tab:modelnet40} in the SO(3)/SO(3) and z/SO(3) modes.

\textit{\textbf{Effect of stage number.}}
To  evaluate the effect of number of stages in PR-invNet, we compare performance of PR-invNet-$i$ with $i=0,1,2,\cdots,5$, and $i$ denotes number of stages excluding initial feature embedding stage. As shown in Fig. \ref{fig:different_hierarchy}, network with 3 stages achieve the highest accuracy. Deeper architectures have larger capacity but marginally decreased performances. This phenomenon was also observed in other graph CNNs~\cite{li2019deepgcns}, and training of deeper networks on point clouds deserves to be more investigated in future work.

\begin{figure}[t]
\centering
\begin{minipage}[t]{0.32\textwidth}
\centering
\includegraphics[width=3.3cm]{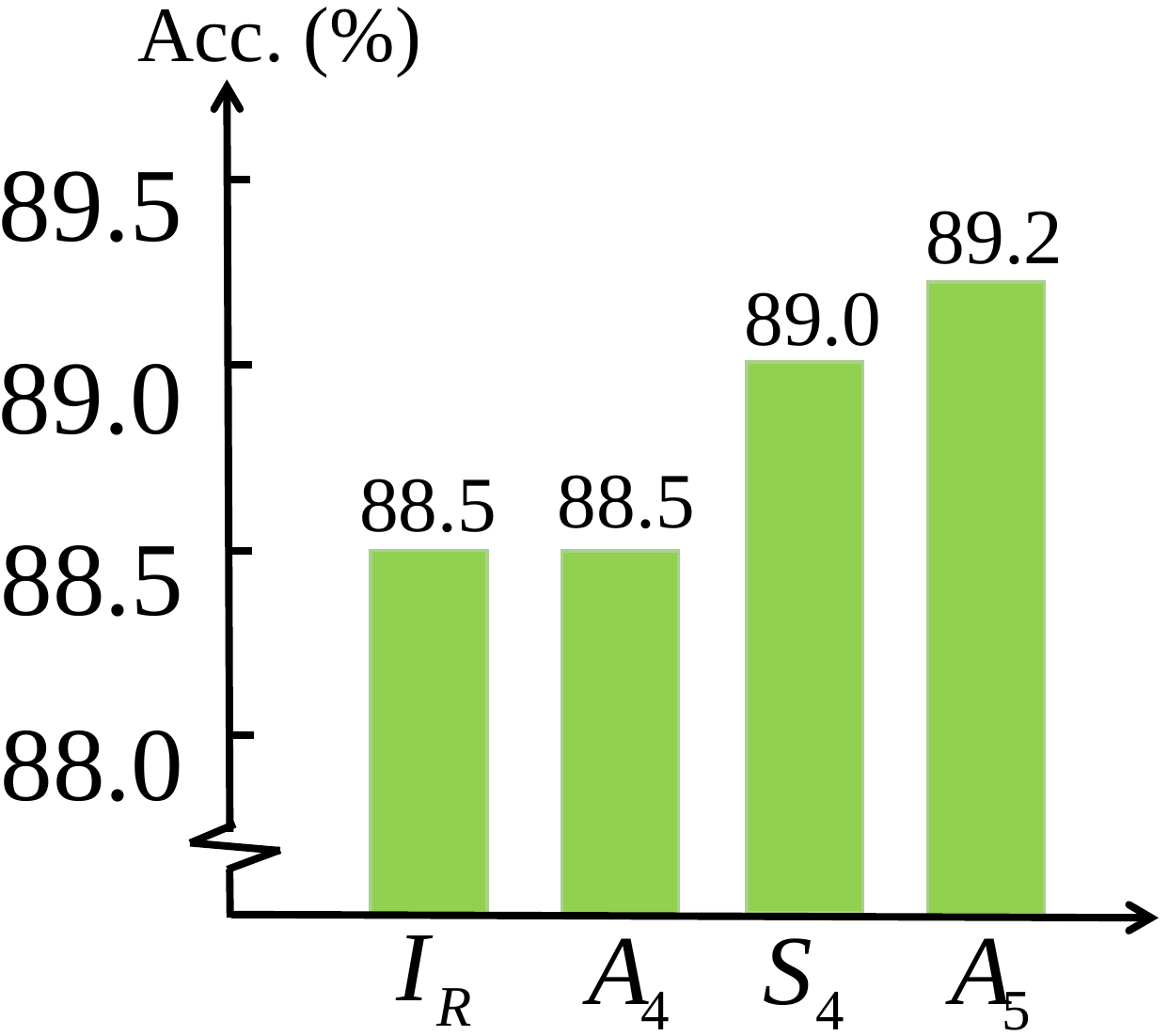}
\caption{Rotation groups.}
\label{fig:different_rotation}
\end{minipage}
\begin{minipage}[t]{.32\textwidth}
\centering
\includegraphics[width=3.3cm]{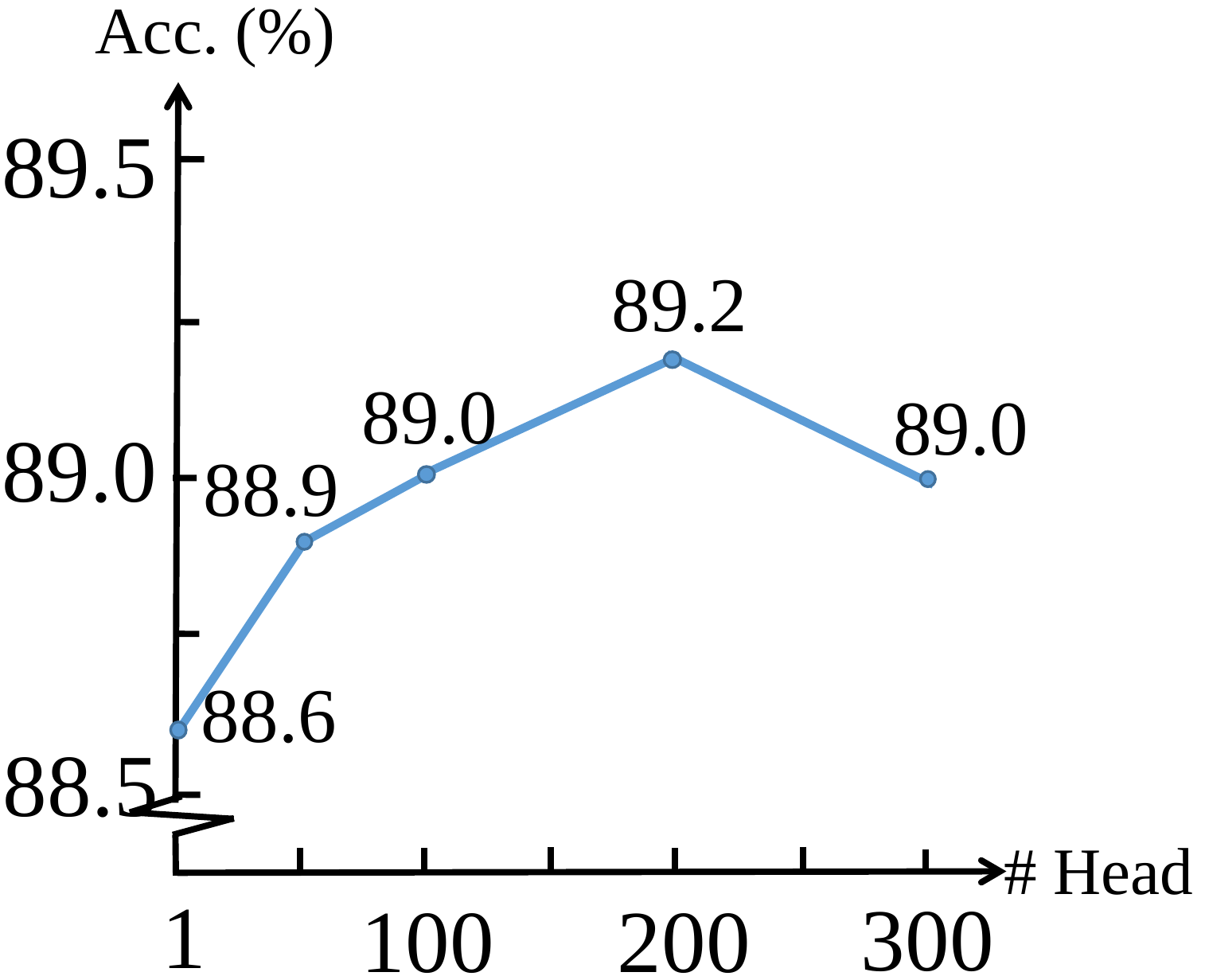}
\caption{Head numbers.}
\label{fig:diff_head}
\end{minipage}
\begin{minipage}[t]{0.32\textwidth}
\centering
\includegraphics[width=3.4cm]{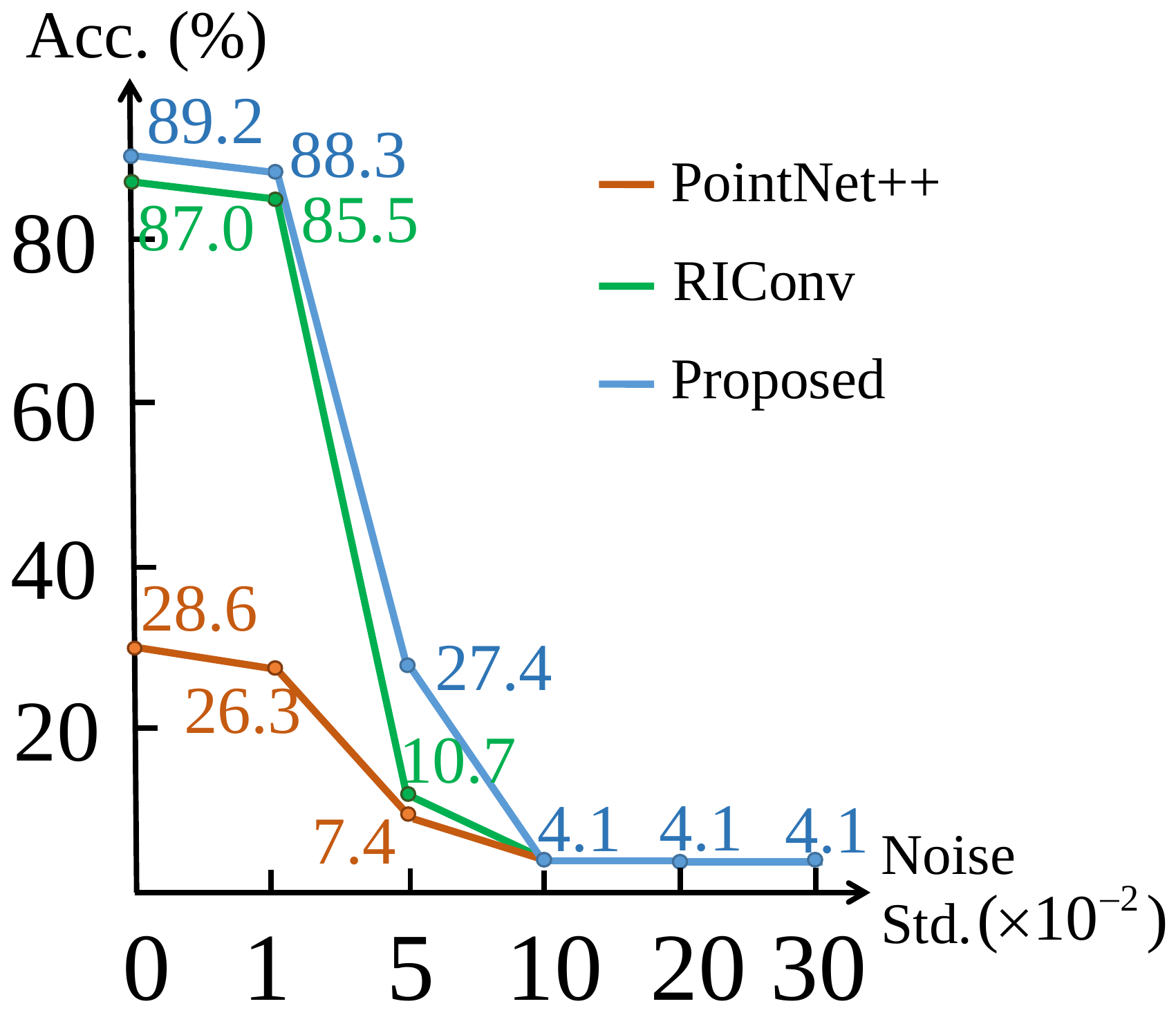}
\caption{Noise levels.}
\label{fig:different_noise}
\end{minipage}
\end{figure}

\textit{\textbf{Design of global shape feature.}} As shown in Fig. \ref{fig:pipleline}, learned features from all PFE-blocks and RFE-blocks are aggregated by MLP for classification. We evaluate PR-invNet-PFE and PR-invNet-RFE that only utilize features from PFE-blocks and RFE-blocks respectively, achieving $88.5\%$, $87.4\%$ classification accuracies on ModelNet40. Result using only features of the last RFE-block is $88.3\%$.
This demonstrates that concatenating all features of PFE-block and RFE-block is more effective.

\textit{\textbf{Effect of PCA-normalization and rot-space.}}
In PR-invNet, we take PCA to normalize shape as network input. We also conduct experiments with networks without PCA-normalization and only with PCA (without using sym-space and rot-space), i.e.,  PR-invNet-noPCA, PR-invNet-PCA. Their classification accuracies in ModelNet40 with z/z mode are $90.0\%$ and $89.1\%$ respectively, but drop to $44.3\%, 74.8\%$ in z/SO(3) mode, which are not rotation-invariant. In SO(3)/SO(3), our pose selector over rot-space upon PCA($89.2\%$) outperforms pure PCA for alignment($88.3\%$). Removing both PCA and pose selector achieves $87.3\%$, showing our pose selection is more effective than PCA.

\textit{\textbf{Effect of rotation group $G_R$.}}
In pose expander of PFE-block, we use the rotation group $G_R$ to construct rot-space, which is taken as alternating group $A_5$ of a regular dodecahedron (or icosahedron). Other alternating (symmetry) group can also be used, such as identity group ($I_R$), or $A_4$, $S_4$ \cite{zimmermann2011finite} respectively corresponding to regular tetrahedron, cube (or octahedron). The classification accuracies using these groups are in Fig. \ref{fig:different_rotation}. Compared with other groups, $A_5$ has the maximum number of elements, resulting in largest rot-space and achieves the highest accuracy. Operating on this largest space, the pose selector has more probability to select a better pose for shape recognition.

\textit{\textbf{Effect of pose selector.}}
In PFE-block, we use a pose selector to select the representative shape pose in rot-space. However, we can also extract features from all shapes from rot-space and then perform max-pooling or average-pooling over them to aggregate features. We compare our pose selector with these aggregation methods and present the classification accuracy and computational cost in Table \ref{tab:different_aggregation}. Since it is infeasible to extract features from all shapes in rot-space containing 120 poses, we compare these variants on the sym-space containing eight shape poses, i.e., rot-space with identity rotation group. We experiment with batch size 12, and as shown in Table \ref{tab:different_aggregation}, our pose selection method achieves higher performance while needing less GPU memory and computational time.

\textit{\textbf{Effect of head number in $\Psi$.}}
For pose selector $\Psi$, we design it as a multi-head neural network, which has 200 heads. We conduct experiments to demonstrate the effect of head numbers. As shown in Fig. \ref{fig:diff_head}, the $\Psi$ with head number as 200 achieves higher performance. Note that $\Psi$ with multi-heads all achieve better accuracies than that with one head.

\begin{table}
\begin{minipage}{0.48\linewidth}
\centering
\caption{Cla. Acc.  on ModelNet40.}
\label{tab:different_aggregation}
\begin{tabular}{l  p{1cm}<{\centering}  p{1cm}<{\centering} p{1cm}<{\centering} }
\hline
\multirow{2}{*}{Method} & Acc. (\%) & Memory (GB) & Time (ms)\\
\hline
Average-pooling & 88.1 & 10.2 & 496.5 \\
Max-pooling &  88.3 &10.2 & 512.3 \\
Proposed & \textbf{88.5} & \textbf{4.4} & \textbf{251.0}\\
\hline
\end{tabular}
\end{minipage}
\begin{minipage}{0.48\linewidth}  
\centering
\caption{Cla. Acc.  on ModelNet40 (\%).}
\label{tab:every_relations}
\begin{tabular}{l  p{0.5cm}<{\centering}  p{0.5cm}<{\centering} p{0.5cm}<{\centering} p{0.5cm}<{\centering} p{0.5cm}<{\centering}}
\hline
~Method      & $C^p_k$  &   $C^g_k$ & $C^f_k$ & $g_k$ & Acc.\\
\hline
~PR-invNet-P &$\times$&$\surd$&$\surd$ & $\surd$& 88.1\\
~PR-invNet-G &$\surd$&$\times$&$\surd$ & $\surd$& 88.8\\
~PR-invNet-F &$\surd$&$\surd$&$\times$& $\surd$& 89.9\\
~PR-invNet-g &$\surd$&$\surd$&$\surd$& $\times$ & 88.5\\
~PR-invNet &$\surd$ &$\surd$ &$\surd$& $\surd$ & \textbf{89.2}\\
\hline
\end{tabular}
\end{minipage}
\end{table}

\textit{\textbf{Design of RFE-block.}}
In RFE-block, we design convolution filters based on three terms, i.e., the relations of point position, geometric feature, and point feature in Eqn. (\ref{eqn:relation}). In Table \ref{tab:every_relations}, we show our networks using only one of these terms, resulting in networks of PR-invNet-P, PR-invNet-G, PR-invNet-F respectively. We also present result of PR-invNet-g only using geometric feature in Eqn. (\ref{eqn:relation}). Their classification accuracies are all inferior compared with full version.
We also conduct experiment that without normalization in Eqn. (\ref{eqn:relation}), achieving classification accuracy as $88.6\%$.

\textit{\textbf{Robustness to noise.}}
We train PR-invNet on ModelNet40 training dataset and test it on test data with various levels of noise under the z/SO(3) mode. We add Gaussian noise with different standard deviation (Std) on each point (coordinates within a unit ball) independently. The overall classification accuracies are in Fig. \ref{fig:different_noise}. PR-invNet keeps robustness under noise with Std of $0.01$. 

\begin{figure}
\begin{center}
\includegraphics[width=0.85\linewidth]{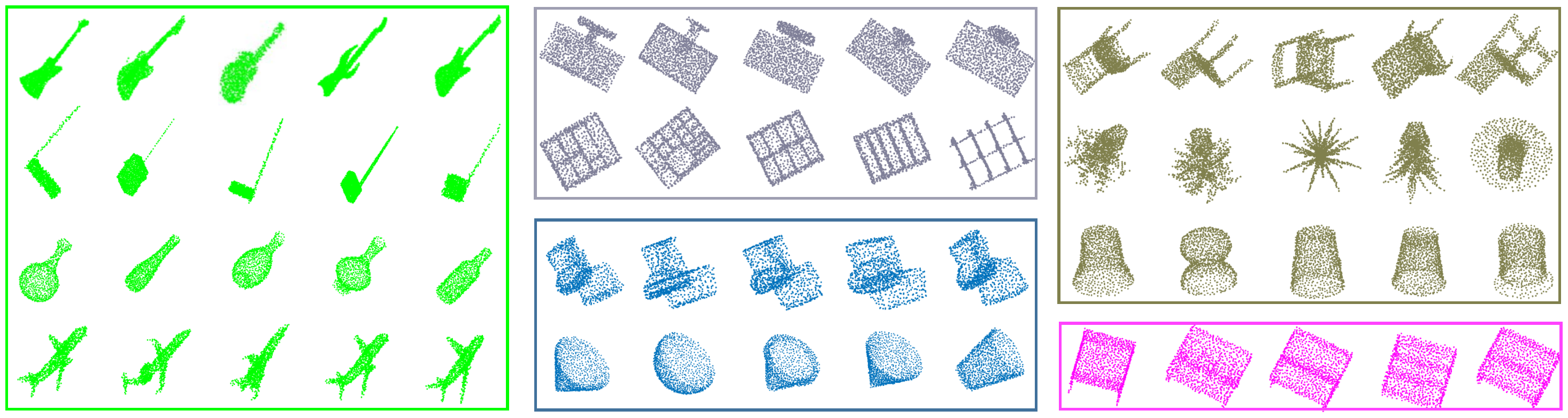}
\end{center}
\caption{Illustration of selected shapes corresponding to different heads in multi-head score network. Shapes in a box highlighted by same color correspond to same head.}
\label{fig:figure_show_centers}
\end{figure}

\textit{\textbf{Visualization for pose selector.}}
In PFE-block, we design multi-head score network $\Psi$ to select a pose from rot-space $H_{SR}(P)$ based on scoring each shape pose by its maximum score belonging to multiple heads. Therefore the selected pose from rot-space corresponds to a certain head. Here we illustrate selected shape poses organized by corresponding heads highlighted in different colors in Fig. \ref{fig:figure_show_centers}. We observe that shapes  corresponding to each head have similar poses. The multi-head score network may enable to learn to align the shapes to clusters of poses using heads in multi-head score network.

\section{Conclusions}
In this work, we focus on rotation-invariant deep network design on point clouds by proposing two rotation-invariant network blocks. The constructed PR-invNet was extensively justified to be effective for rotated 3D shape classification and segmentation. In future work, we are interested to deeply investigate pose alignment capability of PFE-block, and further improve the network architecture.

\textbf{Acknowledgement.} This work was supported by NSFC (11971373, 11690011, U1811461, 61721002) and National Key R\&{D} Program 2018AAA0102201. 

\bibliographystyle{splncs}
\bibliography{refer}
\end{document}